\pdfoutput=1

\documentclass[11pt]{article}
\usepackage[dvipsnames]{xcolor}
\usepackage{acl}

\usepackage{times}
\usepackage{latexsym}

\usepackage[T1]{fontenc}

\usepackage[utf8]{inputenc}

\usepackage{microtype}

\usepackage{booktabs}
\usepackage{rotating}
\usepackage{subfigure}
\usepackage{amsfonts}
\usepackage{amsthm}
\usepackage{bbm}
\usepackage{dsfont}
\usepackage{dcolumn}
\usepackage{caption}
\usepackage{multirow}
\usepackage{amsmath}
\usepackage{amssymb}
\usepackage{mathtools}
\usepackage{makecell}
\usepackage{graphicx}
\usepackage{algorithm}
\usepackage{algpseudocode}

\usepackage{amssymb}%
\usepackage{pifont}%
\newcommand{\cmark}{\ding{51}}%
\newcommand{\xmark}{\ding{55}}%

\newcommand\variablename[1]{\mathop{\textnormal{\slshape #1}}\nolimits}
\newcommand{\DFFOT}{\variablename{DFFOT}}
\newcommand{\DFMIT}{\variablename{DFMIT}}
\newcommand{\SUFF}{\variablename{SUFF}}
\newcommand{\COMP}{\variablename{COMP}}
\newcommand{\CORR}{\variablename{CORR}}
\newcommand{\MONO}{\variablename{MONO}}

\theoremstyle{definition}
\newtheorem{definition}{Definition}[section]
\newtheorem{theorem}{Theorem}[section]

\newtheorem{lemma}[theorem]{Lemma}

\title{A Comparative Study of Faithfulness Metrics for Model Interpretability Methods}

\author{Chun Sik Chan, Huanqi Kong, Guanqing Liang \\
  Wisers AI Lab, Wisers Information Limited \\
  \texttt{\{tonychan, katekong, quincyliang\}@wisers.com}}

\begin{document}
\maketitle
\begin{abstract}

\noindent
Interpretation methods to reveal the internal reasoning processes behind machine learning models have attracted increasing attention in recent years. To quantify the extent to which the identified interpretations truly reflect the intrinsic decision-making mechanisms, various \emph{faithfulness} evaluation metrics have been proposed. However, we find that different faithfulness metrics show conflicting preferences when comparing different interpretations. Motivated by this observation, we aim to conduct a comprehensive and comparative study of the widely adopted faithfulness metrics. In particular, we introduce two assessment dimensions, namely \emph{diagnosticity} and \emph{time complexity}. Diagnosticity refers to the degree to which the faithfulness metric favours relatively faithful interpretations over randomly generated ones, and time complexity is measured by the average number of model forward passes. According to the experimental results, we find that sufficiency and comprehensiveness metrics have higher diagnosticity and lower time complexity than the other faithfulness metrics.
\end{abstract}

\section{Introduction}
\noindent
NLP has made tremendous progress in recent years. However, the increasing complexity of the models makes their behaviour difficult to interpret. To disclose the rationale behind the models, various interpretation methods have been proposed.

Interpretation methods can be broadly classified into two categories: model-based methods and post-hoc methods. Model-based approaches refer to designing simple and white-box machine learning models whose internal decision logic can be easily interpreted, such as linear regression models, decision trees, etc. A post-hoc method is applied after model training and aims to disclose the relationship between feature values and predictions. As pretrained language models \cite{devlin2019bert, liu2019roberta, brown2020language} become more popular, deep learning models are becoming more and more complex. Therefore, post-hoc methods are the only option for model interpretations. Post-hoc interpretation methods can be divided into two categories: gradient-based \cite{simonyan2014deep, 10.5555/3305890.3306024, shrikumar2019learning} and perturbation-based \cite{4407709, zeiler2013visualizing, ribeiro-etal-2016-trust}. Gradient-based methods assume the model is differentiable and attempt to interpret the model outputs through the gradient information. Perturbation-based methods interpret model outputs by perturbing the input data.

To verify whether, and to what
extent, the interpretations reflect the intrinsic reasoning process, various faithfulness metrics have been proposed. Most faithfulness metrics use a removal-based criterion, i.e., removing or retaining only the important tokens identified by the interpretation and observing the changes in model outputs \cite{serrano-smith-2019-attention, chrysostomou-aletras-2021-improving, arras_2017, deyoung-etal-2020-eraser}.

However, we observe that the existing faithfulness metrics are not always consistent with each other and even lead to contradictory conclusions. As shown in the example from our experiments (Table \ref{tab:motivate}), the conclusions that are drawn by two different faithfulness metrics, Sufficiency (SUFF) and Decision Flip - Fraction of Tokens (DFFOT), conflict with each other. More specifically, DFFOT concludes that the interpretation by LIME method is the best among the four interpretations, while SUFF ranks it as the worst. In this case, \emph{which faithfulness metric(s) should we adopt to compare interpretations?}

\begin{table*}
\small
\centering
\begin{tabular}{l|c|c|c|c}
\toprule
\multicolumn{1}{c|}{} & \multicolumn{1}{c|}{} & \multicolumn{2}{c}{\textbf{Faithfulness Metric}} \\ 
\multicolumn{1}{c|}{\multirow{-2}{*}{\textbf{Method}}} & \multicolumn{1}{c|}{\multirow{-2}{*}{\textbf{Interpretation Visualization}}} & \multicolumn{1}{c}{SUFF} & \multicolumn{1}{c}{DFFOT} \\ \midrule

LIME & A cop story that \colorbox{NavyBlue!60}{understands} \colorbox{NavyBlue!10}{the} \colorbox{NavyBlue!35}{medium} amazingly well & \multicolumn{1}{c}{4} & \multicolumn{1}{c}{1} \\ 
Word Omission & \colorbox{NavyBlue!60}{A}\colorbox{NavyBlue!35}{cop} \colorbox{NavyBlue!10}{story} that understands the medium amazingly well & \multicolumn{1}{c}{1} & \multicolumn{1}{c}{4}  \\
Saliency Map & A cop story that understands the \colorbox{NavyBlue!60}{medium} \colorbox{NavyBlue!35}{amazingly} \colorbox{NavyBlue!10}{well} & \multicolumn{1}{c}{3} & \multicolumn{1}{c}{3} \\ 
Integrated Gradients & A cop story that \colorbox{NavyBlue!35}{understands} the \colorbox{NavyBlue!60}{medium} \colorbox{NavyBlue!10}{amazingly} well & \multicolumn{1}{c}{2} & \multicolumn{1}{c}{2}  \\
\bottomrule

\end{tabular}
\caption{An example where different interpretation methods assign different importance scores for the same trained CNN model on SST dataset. The tints of blue mark the magnitude of importance scores for positive sentiment. The numbers 1, 2, 3 and 4 are the rankings of the faithfulness values evaluated by the corresponding faithfulness metrics. Where rank 1 indicates the best, while 4 indicates the worst.}
\label{tab:motivate}
\end{table*}

Motivated by the above observation, we aim to conduct a comprehensive and comparative study of faithfulness metrics. We argue that a good faithfulness metric should be able to effectively and efficiently distinguish between faithful and unfaithful interpretations. To quantitatively assess this capability, we introduce two dimensions, \emph{diagnosticity} and \emph{time complexity}. 

Diagnosticity refers to the extent to which a faithfulness metric prefers faithful rather than unfaithful interpretations. However, due to the opaque nature of deep learning models, it is not easy to obtain the ground truth for faithful interpretation \cite{jacovi-goldberg-2020-towards}. To concretize this issue, we use random interpretations, i.e., randomly assigning importance scores to tokens regardless of the internal processes of the model, as the relatively unfaithful interpretations. In contrast, we treat interpretations generated by interpretation methods as relatively faithful interpretations. In this way, we constructed the hypothesis that a faithfulness metric is diagnostic only if it can clearly distinguish between random interpretations and interpretations generated from interpretation methods. In addition, we introduce time complexity to estimate the computational speed of each metric, by using the average number of model forward passes.

In this paper, we evaluate six commonly adopted faithfulness metrics. We find that the sufficiency and comprehensiveness metrics outperform the other faithfulness metrics, which are more diagnostic and less complex. Secondly, the two correlation-based metrics, namely \textit{Correlation between Importance and Output Probability} and \textit{Monotonicity}, have a promising diagnosticity but fail in terms of the high time complexity. Last but not least, decision flip metrics, such as Fraction of Tokens and Most Informative Token, perform the worst in the assessments.

\noindent
The main contributions of this paper are as follows:
\begin{itemize}
  \setlength\itemsep{0em}
  \item We conduct a comparative study of six widely used faithfulness metrics and identify the inconsistencies issues.
  \item We propose a quantitative approach to assess faithfulness metrics through two perspectives, namely diagnosticity and time complexity.
\end{itemize}

\section{Terminology and Notations} \label{sec:notations}

We first introduce the prerequisite terminology and notations for our discussions.

\paragraph{Terminology} 

A ``classification instance'' is the input and output values of a classification model, which we apply interpretation methods on. An ``interpretation'' of a classification instance is a sequence of scores where each score quantifies the importance of the input token at the corresponding position. An ``interpretation pair'' is a pair of interpretations of the same classification instance.  An ``interpretation method'' is a function that generates an interpretation from a classification instance with the associated classification model.

\paragraph{Notations} 

Let $x$ be the input tokens. Denote the number of tokens of $x$ as $l_x$. Denote the predicted class of $x$ as $c(x)$, and the predicted probability corresponding to class $j$ as $p_j(x)$. 

Assume an interpretation is given. Denote the $k$-th important token as $x_k$. Denote the input sequence containing only the top $k$ (or top $q\%$) important tokens as $x_{:k}$ (or $x_{:q\%}$). Denote the modified input sequence from which a token sub-sequence $x'$ are removed as $x \setminus x'$.

Let $(x,y)$ be a classification instance associated with classification model $m$, and $g$ be an interpretation method. Denote the interpretation of $z$ generated by $g$ as $g(x, y, m)$. Let $u$ be an interpretation, $(u, v)$ be an interpretation pair, and $F$ be a faithfulness metric. Denote the importance score that $u$ assigns to the $i$-th input token as $[u]_i$. Denote the statement ``$u$ is more faithful than $v$'' as ``$u \succ v$'', and the statement ``$F$ considers $u$ as more faithful than $v$'' as ``$u \succ_F v$''.

\section{Faithfulness Metrics} \label{sec:faithfulness_metrics}

An interpretation is called \textit{faithful} if the identified important tokens truly contribute to the decision making process of the model. Mainstream faithfulness metrics are removal-based metrics, which measure the changes in model outputs after removing important tokens. 

We compare the most widely adopted faithfulness metrics, introduced as follows.

\paragraph{Decision Flip - Most Informative Token (DFMIT)} Introduced by \citet{chrysostomou-aletras-2021-improving}, this metric focuses on only the most important token. It assumes that the interpretation is faithful only if the prediction label is changed after removing the most important token, i.e.
\begin{equation*}
\label{eq:dfmit}
  \DFMIT =
    \begin{cases}
      1 & \text{if $c(x) \neq c(x \setminus x_{:1}))$}\\
      0 & \text{if $c(x) = c(x \setminus x_{:1}))$}\\
    \end{cases}       
\end{equation*}
A score of 1 implies that the interpretation is faithful.

\paragraph{Decision Flip - Fraction of Tokens (DFFOT)} This metric measures faithfulness as the minimum fraction of important tokens needed to be erased in order to change the model decision \cite{serrano-smith-2019-attention}, i.e.
\begin{equation*}
\label{eq:dffot}
  \DFFOT =
    \begin{cases}
      \min \frac{k}{l_x} & \text{s.t. $c(x) \neq c(x \setminus x_{:k})$}\\
      1 & \text{if $c(x) = c(x \setminus x_{:k})$ for any k}\\
    \end{cases}       
\end{equation*}
 If the predicted class change never occurs even if all tokens are deleted, then the score will be 1. A lower value of DFFOT means the interpretation is more faithful. 

\paragraph{Comprehensiveness (COMP)} As proposed by \citet{deyoung-etal-2020-eraser}, comprehensiveness assumes that an interpretation is faithful if the important tokens are broadly representative of the entire input sequence. It measures the faithfulness score by the change in the output probability of the original predicted class after the important tokens are removed, i.e.
\begin{equation*}
\label{eq:comp}
    \COMP = \frac{1}{|B|} \sum_{q \in B}(p_{c(x)}(x) - p_{c(x)}(x \setminus x_{:q\%}))
\end{equation*}
We use $q \in B = \{1, 5, 10, 20, 50\}$ as in the original paper. A higher comprehensiveness score implies a more faithful interpretation.

\paragraph{Sufficiency (SUFF)} Also proposed by \citet{deyoung-etal-2020-eraser}, this metric measures whether the important tokens contain sufficient information to retain the prediction. It keeps only the important tokens and calculates the change in output probability compared to the original specific predicted class, i.e.
\begin{equation*}
    \SUFF = \frac{1}{|B|} \sum_{q \in B}(p_{c(x)}(x) - p_{c(x)}(x_{:q\%}))
\end{equation*}
 We use $q \in B = \{1, 5, 10, 20, 50\}$ as in the original paper. The lower the value of SUFF means that the interpretation is more faithful.

\paragraph{Correlation between Importance and Output Probability (CORR)}  This metric assumes that the interpretation is faithful if the importance of the token and the corresponding predicted probability when the most important token is continuously removed is positively correlated \cite{arya2019explanation}, i.e. 
\begin{equation*}
\label{eq:corr}
    \CORR = - \rho (\boldsymbol{u}, \boldsymbol{p})
\end{equation*}
where $\boldsymbol{u}$ denotes the token importance in descending order and $\boldsymbol{p} = [p_{c(x)}(x \setminus x_1), p_{c(x)}(x \setminus x_2),  ... , p_{c(x)}(x \setminus x_{l_x})]$. $\rho(\cdot)$ denotes the Pearsons correlation. The higher the correlation the more faithful the interpretation is.

\paragraph{Monotonicity (MONO)} This metric assumes that an interpretation is faithful if the probability of the predicted class monotonically increases when incrementally adding more important tokens \cite{arya2019explanation}. Starting from an empty vector, the features are gradually added in ascending order of importance, and the corresponding classification probabilities are noted. Monotonicity is calculated as the correlation between the feature importance and the probability after adding the feature, i.e.
\begin{equation*}
\label{eq:mono}
    \MONO = \rho (\boldsymbol{u}, \boldsymbol{p})
\end{equation*}
where $\boldsymbol{u}$ denotes the token importance in descending order and $\boldsymbol{p} = [p_{c(x)}(x), p_{c(x)}(x \setminus x_{:1}), p_{c(x)}(x \setminus x_{:2}),  ... , p_{c(x)}(x \setminus x_{:(l_x-1)})]$. $\rho(\cdot)$ denotes the Pearsons correlation. The higher the monotonicity the more faithful the interpretation is.

\section{Evaluation of Faithfulness Metrics}

In this section, we propose an evaluation paradigm for faithfulness metrics by addressing two aspects: (1) diagnosticity and (2) time complexity. They are the two complementary and important factors in selecting a faithfulness metric for assessing the faithfulness of interpretations.

\subsection{Diagnosticity of Faithfulness Metric} 
As we have observed in Table \ref{tab:motivate}, faithfulness metrics might disagree with each other on faithfulness assessment. This naturally raises a question: \textit{Which faithfulness metric(s) should we trust?}

To the best of our knowledge, there is no preceding work in quantifying the effectiveness of faithfulness metrics. As a first attempt, we introduce \textit{diagnositicity}, which is intended to measure ``the degree to which a faithfulness metric favours faithful interpretations over unfaithful interpretations''. Intuitively, the higher the diagnosticity the more effective the faithfulness metric is.

\subsubsection{Definition of Diagnosticity} 

\begin{definition}[\textbf{Diagnosticity}]
\label{def:diag}
We define the diagnosticity of a faithfulness metric as the probability that given an interpretation pair $(u, v)$ such that $u$ is more faithful than $v$, the faithfulness metric also considers $u$ as more faithful than $v$, i.e.
\[\text{D}(F) = \text{P}(u \succ_F v| u \succ v)\]
\end{definition}

As we will see later in this section, a set of interpretation pairs $(u, v)$ such that $u \succ v$  is required for estimating diagnosticity. Constructing such a dataset leads us to a paradox: we cannot be \textit{guaranteed} that some generated interpretation is more faithful than the others when the measurement of faithfulness is still under debate. It is more realistic to assume that we can generate an interpretation pair $(u, v)$ such that $u$ is \textit{very likely} to be more faithful than $v$. Thus, we relax the condition in Definition \ref{def:diag} to a probabilistic one as follows.

\begin{definition}[\textbf{$\varepsilon$-diagnosticity}]
\label{def:approx_diag}
Let $(u, v)$ be any interpretation pair, and $0 \leq \varepsilon \leq 1$. The $\varepsilon$-diagnosticity of a faithfulness metric $F$ is defined as 
\[\text{D}_{\varepsilon}(F) = \text{P}(u \succ_F v| \text{P} (u \succ v) > 1 - \varepsilon)\] 
\end{definition}

In the above definition, $\varepsilon$ represents the uncertainty in comparing the faithfulness of $u$ and $v$. In the next Theorem, we show that $\varepsilon$-diagnosticity effectively approximates diagnosticity as long as $\varepsilon$ is small enough. 

\begin{theorem}[\textbf{Error Bound of $\varepsilon$-diagnosticity}]
\label{thm:approx_diag}
We can approximate diagnosticity with $\varepsilon$-diagnosticity with error less than $\varepsilon$, i.e. 
\[
| \text{D}_\varepsilon (F) - \text{D} (F) | < \varepsilon
\]
\end{theorem}

\noindent
The proof is provided in Appendix
\ref{sec:appendix_proof1}.

\subsubsection{Estimation of Diagnosticity} 
In the following, we show how we estimate $\varepsilon$-diagnosticity with a set of interpretation pairs $(u, v)$ where the $u$ is \textit{very likely} to be more faithful than $v$, namely an \textit{$\varepsilon$-faithfulness golden set} where $\varepsilon$ is small. 

\begin{definition}[\textbf{$\varepsilon$-faithfulness golden set}]
\label{def:fgs}
Let $0 \leq \varepsilon \leq 1$. A set $Z_{\varepsilon}$ of interpretation pairs is called a $\varepsilon$-faithfulness golden set, if it satisfies the following conditions. 
\begin{enumerate}
    \item All interpretation pairs in $Z_\varepsilon$ are independent and identically distributed (i.i.d.).
    \item $\text{P}(u \succ v) > 1 - \varepsilon $ for any interpretation pair $(u, v) \in Z_\varepsilon$.
\end{enumerate}
\end{definition}

\begin{lemma}
\label{lm:expected_value}
Let $\mathbbm{1} (\cdot)$ be the indicator function which takes a value 1 when the input statement is true and a value 0 when it is false. Then $\mathbbm{1} (u \succ_F v) | (\text{P} (u \succ v) > 1 - \varepsilon)$ is a random variable and its expected value is equal to $\varepsilon$-diagnosticity, i.e.
\[
\text{D}_{\varepsilon}(F) = \mathop{\mathbb{E}} [\mathbbm{1} (u \succ_F v) | \text{P} (u \succ v) > 1 - \varepsilon]
\]
\end{lemma}
\noindent
The proof is provided in Appendix
\ref{sec:appendix_proof2}.

As a result, given an $\varepsilon$-faithfulness golden set $Z_{\varepsilon}$, we can estimate the $\varepsilon$-diagnosticity of a faithfulness metric $F$ by estimating the expected value in Lemma \ref{lm:expected_value}. Then by the law of large numbers, we can simply estimate the expected value by computing the average value of $\mathbbm{1} (u \succ_F v)$ on $Z_\varepsilon$, i.e.
\begin{equation}
\label{eqn:diag_est}
    \text{D}_{\varepsilon}(F) \approx \frac{1}{|Z_\varepsilon|} \sum_{(u, v) \in Z_\varepsilon} \mathbbm{1} (u \succ_F v)
\end{equation}

When $|Z_\varepsilon|$ is large enough, we will have $|\frac{1}{|Z_\varepsilon|} \sum_{(u, v) \in Z_\varepsilon} \mathbbm{1} (u \succ_F v) - D(F)| < \varepsilon$ according to Theorem \ref{thm:approx_diag}. 

\subsubsection{Generation of an $\varepsilon$-faithfulness golden set} 
\label{sec:fgs}

According to Theorem \ref{thm:approx_diag} and Lemma \ref{lm:expected_value}, we can estimate the diagnosticity of any faithfulness metric using Equation \ref{eqn:diag_est} as long as we have an $\varepsilon$-faithfulness golden set where $\varepsilon$ is small enough. 

We called the $u$ and $v$ in Definition \ref{def:fgs} a \textbf{relatively faithful interpretation} and a \textbf{relatively unfaithful interpretation} respectively. Next, we discuss the processes to generate them respectively.

\paragraph{Generating Relatively Unfaithful Interpretations}
By definition, a faithful interpretation is an interpretation that truly reflects the underlying decision making process of the classification model. Therefore, an unfaithful interpretation is one that is completely irrelevant to the underlying decision making process of the classification model. We propose to generate relatively unfaithful interpretations by assigning a random importance score to each token in the input sequence, i.e. $[v]_i \sim \text{Uniform}(0, 1)$ for any token $1 \leq i \leq l$, where Uniform denotes the uniform distribution.

\paragraph{Generating Relatively Faithful Interpretations}
We propose to generate relatively faithful interpretations with the interpretation methods that infer interpretations from the underlying mechanism of the classification model. There are two mainstream categories of interpretations methods that satisfy this requirement \cite{alvarezmelis2018robustness}:

\begin{itemize}
    \item \textbf{Perturbation-based}: Relying on querying the model around the classification instance to infer the importance of input features.
    \item \textbf{Gradient-based}: Using information from gradients to infer the importance of input features.
\end{itemize}

We select the representative methods from both categories and introduce them in the following. 

\begin{itemize}
    \item \textbf{Perturbation-based - LIME } \cite{ribeiro-etal-2016-trust}: For each classification instance, a linear model on the input space is trained to approximate the local decision boundary, so that the learned coefficients can be used to quantify the importance of the corresponding input features on the model prediction.
    
    \item \textbf{Perturbation-based - Word Omission (WO) } \cite{4407709}: For each $i$-th input token, WO quantifies the importance of the input token by the change in output probability after removing it from the original input sequence, i.e. $p_{c(x)}(x) - p_{c(x)}(x_{\setminus \{i\}})$.
    
    \item \textbf{Gradient-based - Saliency Map (SA)}  \cite{simonyan2014deep}: For each $i$-th input token, SA computes the gradients of the original model output with respect to the embedding associated with the input token, i.e. $\frac{\partial p_{c(x)} (z)}{\partial e(z)_i} |_{z=x}$, and quantifies the importance of the input token by taking either the mean or the $l2$ norm of the gradients in the embedding dimension. We denote the former approach as SA$_\mu$ and the later approach as SA$_{l2}$
    
    \item \textbf{Gradient-based - Integrated Gradients (IG)}  \cite{simonyan2014deep}: As shown by \citet{simonyan2014deep}, Integrated Gradients provides more robust interpretations than Saliency Map in general. For each $i$-th input token, it approximates the integral of the gradients of the original model output with respect to the embedding corresponding to the input token along a straight line from a reference point $x_0$ to the original input sequence, i.e. $\int_{x_0 \rightarrow x} \frac{\partial p_{c(x)} (z)}{\partial e(z)_i} \,dz$ , and quantifies the importance of the input token by taking either the mean or the $l2$ norm of the integral in the embedding dimension. We denote the former approach as IG$_\mu$ and the later approach as IG$_{l2}$.

\end{itemize}

\begin{algorithm}[t!]
  \caption{An $\varepsilon$-faithfulness golden set generation mechanism.}
  \label{alg:gen}
  \begin{algorithmic}
    \Require 
      $X$: A set of i.i.d. classification instances associated with classification model $m$; \\
      $G$: The set of interpretation methods for generating relatively faithful interpretations, i.e. \{LIME, WO, SA$_{\mu}$, SA$_{l2}$, IG$_{\mu}$, IG$_{l2}$\}; \\
      $K$: Sample size; 
    \Ensure
      An $\varepsilon$-faithfulness golden set $Z$;
    \State $Z \leftarrow \{\}$;
    \State For 1 to $K$;
    \State \qquad $(x, y) \leftarrow \text{RandomSampler}(X)$;
    \State \qquad $g \leftarrow \text{RandomSampler}(G)$;
    \State \qquad $u \leftarrow g (x, y, m)$
    \State \qquad $v \leftarrow r \in \mathbb{R}^{l_x}$ where $[r]_i \sim \text{Uniform}(0, 1)$;
    \State \qquad $Z \leftarrow Z \cup \{(u, v)\}$;\\
    \Return $Z$;
  \end{algorithmic}
\end{algorithm}

The interpretations generated using the above interpretation methods are \textit{highly likely} to be more faithful than the randomly generated interpretations because the generation processes of the former ones actually involve inferences from model behaviours, while the random generation process is independent of any model behaviour. Therefore, in principle, the set of generated interpretation pairs will have a small value of  $\varepsilon$ in Definition \ref{def:fgs}.

In Algorithm \ref{alg:gen}, we propose a mechanism to generate an $\varepsilon$-faithfulness golden set from a set of i.i.d. classification instances based on the above processes. Note that the generated interpretation pairs will satisfy the first condition in Definition \ref{def:fgs} because they are generated from i.i.d. samples, and will satisfy the second condition in Definition \ref{def:fgs} with a presumably small $\varepsilon$ as we have discussed.

\subsection{Time Complexity of Faithfulness Metric} 

Two of the main applications of faithfulness metrics are  (1) evaluating interpretation methods based on their average faithfulness scores on a dataset; and (2) gauging the quality of individual interpretations by spotting out ``unfaithful'' interpretations. 

Time complexity is an important aspect in evaluating faithfulness metrics because a fast faithfulness metric will shorten the feedback loop in developing faithful interpretation methods, and would allow runtime faithfulness checking of individual interpretations in a production environment.

\paragraph{Measurement of time complexity} 

From the definitions of the faithfulness metrics in Section \ref{sec:faithfulness_metrics}, we observe that their computations are dominated by model forward passes, which are denoted as $c(\cdot)$ or $p(\cdot)$. Thus, we measure the time complexities of the faithfulness metrics in \textit{number of model forward passes}.

\section{Experimental Setup \footnote{Code will be available at \url{https://github.com/Wisers-AI/faithfulness-metrics-eval}}} 

\paragraph{Datasets} We conduct experiments on three text classification datasets used in \cite{wiegreffe-pinter-2019-attention}: (i) Stanford Sentiment Treebank (\textbf{SST}) \cite{socher-etal-2013-recursive}; (ii) IMDB Large Movie Reviews (\textbf{IMDB}) \cite{maas-etal-2011-learning}; (iii) AG News Corpus (\textbf{AG}) \cite{10.5555/2969239.2969312}. We summarize the dataset statistics in Table \ref{table:dataset}.

\begin{table}[t!]
\small
\centering
\begin{tabular}{l | c c c}
\toprule
\textbf{Dataset} & \textbf{Splits (Train / Test)} & \multicolumn{2}{c}{\textbf{Model perf. (F1)}} \\
 & & BERT & CNN \\
\midrule
SST & 6,920 / 1,821 & .917 & .804 \\
IMDB & 25,000 / 25,000 & .918 & .864 \\
AG & 120,000 / 7,600 & .946 & .919 \\
\bottomrule
\end{tabular}
\caption{Dataset statistics and model performances (Macro-F1) on test sets.}
\label{table:dataset}
\end{table}

\paragraph{Text classification models}
\label{subsection:models}

\noindent
We adopt two most common model architectures for text classification: (i) \textbf{BERT} \cite{devlin-etal-2019-bert}; (ii) \textbf{CNN} \cite{kim-2014-convolutional}. The former one encodes contextualized representations of tokens and has higher accuracy in general, but at a cost of consuming more memory and computational resources. The latter one uses pretrained word embeddings as token representations and is lighter and faster. Their performances on test data sets are shown in Table \ref{table:dataset}. The implementation details of both models can be found in Appendix \ref{sec:appendix_model}. 

\paragraph{$\varepsilon$-faithfulness golden set}

\noindent
For each dataset and text classification model, we transform the test set into a set of classification instances and feed it into Algorithm \ref{alg:gen} to generate an $\varepsilon$-faithfulness golden set with a size of 8,000 ($K$ in Algorithm \ref{alg:gen}). The implementation details of interpretation methods can be found in Appendix \ref{sec:appendix_interpret}.

\section{Results and Discussion}

\paragraph{Diagnosticity}

\begin{table}[t!]
\small
\begin{tabular}{l | rrrr}
\toprule
\textbf{Faithfulness} &   \multicolumn{4}{c}{\textbf{Diagnosticity (\%)}}\\
\textbf{metric} &     SST &   IMDB &     AG &  Average \\
\midrule
 & \multicolumn{4}{c}{\textbf{BERT}} \\
DFMIT  &  14.79  &  6.07  &  3.34  & 8.07 \\
DFFOT  &  65.16 &  72.02 &  65.68 &    67.62 \\
SUFF   &  71.03 &  79.33 &  70.42 &    73.60 \\
COMP   &  75.38 &  \underline{\textbf{80.44}}   & \underline{\textbf{74.23}}   &   \underline{\textbf{76.69}}  \\
CORR   &  65.46 &  68.06 &  67.23 &    66.91 \\
MONO   & \underline{\textbf{75.87}}  &  75.82 &  68.33 &    73.34 \\
\midrule
 & \multicolumn{4}{c}{\textbf{CNN}} \\
DFMIT  &  17.29 &    9.27 &   4.84  &   10.47 \\
DFFOT  &  63.76 &  70.74 &  57.61 &    64.04 \\
SUFF   &  71.54 &  75.91 &  77.97 &    75.14 \\
COMP   &  71.39 &  73.46 &  \underline{\textbf{81.73}} &    \underline{\textbf{75.53}} \\
CORR   &  72.17 &  68.92 &  71.82 &    70.97 \\
MONO   &   \underline{\textbf{72.39}} &   \underline{\textbf{77.09}} &  75.12 &    74.87 \\
\bottomrule
\end{tabular}
\caption{Diagnosticities of all faithfulness metrics on all datasets for both BERT and CNN models. The rightmost column states the average diagnosticities over three datasets. In each column, we underline the highest value.}
\label{table:diagnosticity}
\end{table}
  
We estimate the diagnositicities of the faithfulness metrics in Section \ref{sec:faithfulness_metrics} on all datasets for both CNN and BERT models. The results are shown in Table \ref{table:diagnosticity}.

COMP and SUFF have the highest and the second highest average diagnosticites for both models. Hence, they are the most effective faithfulness metrics. We also observe that COMP has higher diagnosticities than SUFF on all datasets for BERT model. This can be explained by the contextualization property of Transformer encoders \cite{NIPS2017_3f5ee243}: the hidden state of each token depends on all other tokens in the input sequence. Removing a portion of the important tokens will alter the whole context, and is likely to cause a dramatic change in model output. 

DFMIT and DFFOT have the lowest and the second lowest average diagnosticities. Removing the most important token is usually not creating enough perturbation to flip the original model decision. In fact, the probability of decision flipping by removing the most important token is $\leq 14\%$ for recent state-of-the-art interpretation methods \cite{chrysostomou-aletras-2021-improving}. As a result, up to 86\% of interpretations are considered as indifferent by DFMIT. For DFFOT, the probability of decision flipping by removing the important tokens in order does not only depend on the quality of interpretation but also depends on any model bias towards certain classes. For instance, decision flipping will be less likely to occur if the predicted class on the original input is the same as the one on the empty input sequence. Therefore, we found that decision flipping metrics (DFMIT, DFFOT) are less effective than the metrics that operate on output probabilities (SUFF, COMP, CORR, MONO).

\paragraph{Time complexity}

We compare the time complexities of the faithfulness metrics in Section \ref{sec:faithfulness_metrics} measured in number of model forward passes. We first analyze their time complexities based on their definitions in Table \ref{table:complexity_analyze} and then measure their actual time complexities on all datasets in Table \ref{table:complexity}. Note that the time complexity here is equal to the number of model forward passes.

DFMIT is the fastest faithfulness metric, which requires only one model forward pass. DFFOT has a non-deterministic time complexity, which depends on how fast the decision flipping occurs, and it is the second slowest faithfulness metrics on all datasets. SUFF and COMP are the second fastest faithfulness metric on average, which require at most 5 model forward passes. CORR and MONO are the slowest faithfulness metrics, which have time complexity equal to the number of input tokens.

\begin{table}[t!]
\small
\centering
\begin{tabular}{ l | c c }
\toprule
\textbf{Faithfulness} &  \multicolumn{2}{c}{\textbf{Time complexity - Analysis}} \\
\textbf{metric} &  \multicolumn{2}{c}{\textbf{(\#(model forward passes))}} \\
 & Deterministic & Value or range \\
\midrule
DFMIT  & \cmark & 1 \\
DFFOT  & \xmark & $[1,l_x]$ \\
SUFF  & \cmark & $min(5, l_x)$ \\
COMP  & \cmark & $min(5, l_x)$ \\
CORR  & \cmark & $l_x$ \\
MONO  & \cmark & $l_x$ \\
\bottomrule
\end{tabular}
\caption{Analysis of the time complexities of faithfulness metrics. $l_x$ denotes the number of input tokens.}
\label{table:complexity_analyze}
\end{table}

\begin{table}[t!]
\small
\centering
\begin{tabular}{ l | c c c c}
\toprule
\textbf{Faithfulness} &  \multicolumn{4}{c}{\textbf{Time complexity - Actual}} \\
\textbf{metric} &  \multicolumn{4}{c}{\textbf{(\#(model forward passes))}} \\
 & SST & IMDB & AG & Average \\
\midrule
DFMIT  & 1.0 & 1.0 & 1.00 & 1.0 \\
DFFOT  & 9.3 & 78.7 & 30.0 & 39.4 \\
SUFF & 5.0 & 5.0 & 5.0 & 5.0 \\
COMP & 5.0 & 5.0 & 5.0 & 5.0 \\
CORR & 20.3 & 193.1 & 47.7 & 87.1 \\
MONO & 20.3 & 193.1 & 47.7 & 87.1 \\
\bottomrule
\end{tabular}
\caption{Actual time complexities of faithfulness metrics measured by the average number of model passes on each dataset.}
\label{table:complexity}
\end{table}

\paragraph{Which faithfulness metric(s) should we adopt?}

\begin{figure}[t!]
\centering
    \includegraphics[width=1\linewidth]{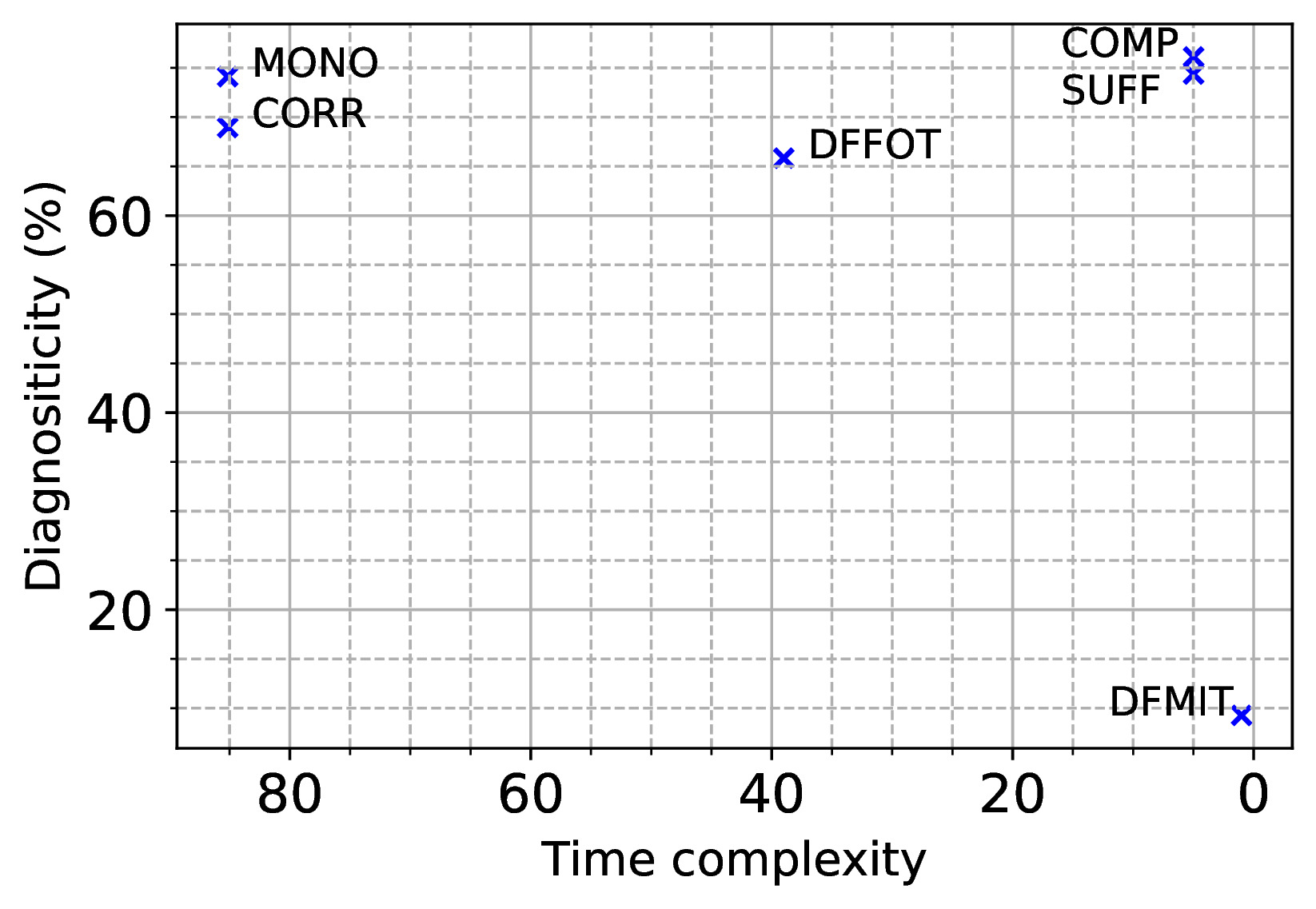}
    \caption{Diagnosticity vs time complexity for faithfulness metrics. The values are averages over all datasets and classification models. The faithfulness metrics near the top-right corner are more desirable than those near the bottom-left corner.} 
    \label{fig:diag_vs_comp}
    
\end{figure}

In Figure \ref{fig:diag_vs_comp}, we evaluate the faithfulness metrics by both their diagnosticities and time complexities.

Figure \ref{fig:diag_vs_comp} suggests that we should always adopt COMP and SUFF. Because (i) they have higher diagnosticities and lower time complexities than DFFOT, ; (ii) they have a similar level of diagnosticity and much lower time complexities than CORR and MONO; (iii) DFMIT has diagnosticity less than 0.1, which is below an acceptable level. We would prefer COMP and SUFF over DFMIT even though it has the lowest time complexity.

Note that our evaluation framework can be used to compare any faithfulness metrics. In general, we prefer faithfulness metrics that have higher diagnosticities and lower time complexities, i.e. closer to the top-right corner in Figure \ref{fig:diag_vs_comp}. But what if one has a higher diagnosticity and the other one has a lower time complexity? In this case, we should consider diagnosticity first: a faithfulness metric should not be used if it cannot effectively assess faithfulness, i.e. diagnosticity below a certain threshold. In scenarios where we are subject to constraints of hardware or timeliness, we might need to select a faster metric with a lower but acceptable level of diagnosticity.

\section{Related Work}
\paragraph{Interpretation methods} Interpretation methods can be roughly classified into two categories: model-based methods and post-hoc methods. Model-based methods refer to the construction of simple machine learning models whose internal decision logic can be easily interpreted, such as linear regression models, decision trees, etc. Post-hoc methods interpret the internal reasoning process behind the model after training. Generally, post-hoc methods can be divided into gradient-based and perturbation-based. A gradient-based interpretation method assumes deep learning model is differentiable and discloses the decision making mechanism of the model according to the gradient information \cite{simonyan2014deep, 10.5555/3305890.3306024, shrikumar2019learning}. A perturbation-based interpretation method interprets the model by perturbing the input of data samples and measuring how the predictions change \cite{4407709, zeiler2013visualizing, ribeiro-etal-2016-trust}.

\paragraph{Interpretation method evaluation} To assess the quality of different interpretation methods, various evaluation metrics have been proposed. Existing evaluation methods on interpretations can be broadly classified into two categories, plausibility and faithfulness. Plausibility measures if the interpretation agrees with human judgments on how a model makes a decision \cite{ribeiro-etal-2016-trust, doshivelez2017rigorous, lundberg2017unified, deyoung-etal-2020-eraser}. However, even if the 
interpretation conforms to human criteria, it is not certain that it truly reflects the underlying decision mechanism behind the model. To this end, faithfulness measures the extent to which the inner decision-making mechanism actually relies on the identified important features \cite{arras_2017, serrano-smith-2019-attention, jain-wallace-2019-attention, wiegreffe-pinter-2019-attention, deyoung-etal-2020-eraser, chrysostomou-aletras-2021-improving}. 

In general, existing faithfulness metrics are developed through a removal-based criterion, which measures the changes in model output when perturbing or removing tokens identified as important by the interpretation. \citet{serrano-smith-2019-attention} proposed a decision flipping metric that evaluates the proportion of tokens that need to be erased in order to change the model decision. Also using decision flip as an indicator, \citet{chrysostomou-aletras-2021-improving} introduces a metric that counts the average flips that occur when removing the most important token marked by the interpretation method. In addition to decision flips, changes in model output probabilities by removing or retaining important tokens is also widely used to measure faithfulness \cite{arras_2017, arya2019explanation, deyoung-etal-2020-eraser}.
                          
Some recent work also focuses on the study of faithfulness metrics. \citet{jacovi-goldberg-2020-towards} argued that the definition of faithfulness remains inconsistent and informal, and provided concrete guidelines on how evaluations of interpretation methods should and should not be conducted. More recently, \citet{yin2021faithfulness} discussed the limitations of removal-based faithfulness metrics and proposed two other quantitative criteria, namely sensitivity and stability. Different from the aforementioned previous work that does not focus on assessing faithfulness metrics, we mainly focus on the measurement of faithfulness and conduct a comprehensive study of existing faithfulness metrics.

\section{Conclusion}
In this paper, we propose a framework to quantitatively evaluate six widely adopted faithfulness metrics in terms of diagnosticity and time complexity. In particular, diagnosticity measures whether the faithfulness metric correctly favours relatively faithful interpretations over random ones; time complexity is concerned with computational efficiency, estimated by the average number of model forward passes. The experimental results show that sufficiency and comprehensiveness metrics outperform the other faithfulness metrics with higher diagnosticity and lower time complexity. For this reason, we suggest using these two metrics for faithfulness evaluation. We hope our work will bring more awareness to the standardization of faithfulness measurement. For future work, we would like to explore evaluating faithfulness metrics using a white-box model such as linear regression, from which we can derive an intrinsically faithful interpretation as the ``ground truth''.

\bibliography{anthology,custom}
\bibliographystyle{acl_natbib}

\appendix
\setcounter{section}{0}

\section{Proof of Theorem \ref{thm:approx_diag}}
\label{sec:appendix_proof1}

\begin{proof}
\small
Let $(u, v)$ be an interpretation pair. Then
\[
\begin{split}
& \text{P}(u \succ_F v|\text{P} (u \succ v) = 1 - \varepsilon) \\ 
& = \text{P}(u \succ_F v | u \succ v)(1-\varepsilon) + \text{P}(u \succ_F v | u \nsucc v) \varepsilon \\
& = \text{D}(F) + [\text{P}(u \succ_F v | u \nsucc v)-\text{P}(u \succ_F v | u \succ v)] \varepsilon
\end{split}
\]
Since $-1 \leq \text{P}(u \succ_F v | u \nsucc v)-\text{P}(u \succ_F v | u \succ v) \leq 1$, we have
\[
|\text{P}(u \succ_F v|\text{P} (u \succ v) = 1 - \varepsilon) - \text{D}(F)| \leq \varepsilon
\]
\end{proof}

\section{Proof of Lemma \ref{lm:expected_value}}
\label{sec:appendix_proof2}

\begin{proof}
From Definition \ref{def:approx_diag}, we have $\mathbbm{1} (u \succ_F v) | (\text{P} (u \succ v)  > 1 - \varepsilon) \sim \text{Bernoulli}(p)$, where $p = \text{D}(F)$. Then based on the property of Bernoulli distribution, we know that the expected value of the random variable is equal to $p$.
\end{proof}

\section{Implementation Details}

\subsection{Text classification models}
\label{sec:appendix_model}
The text classification models are all implemented in PyTorch \footnote{https://pytorch.org/}. For BERT, we use the ``bert-base-uncased'' from Huggingface transformers  \footnote{https://github.com/huggingface/transformers} as the pretrained model . We use the same set of hyperparameters regardless of dataset for fine-tuning: dropout rate 0.2, AdamW \cite{loshchilov2019decoupled} with an initial learning rate 2e-5, batch size 32 with no warmup steps. We set the maximum number of finetuning epochs to be 10 and perform early stopping when the performance
on the test set does not improve for 3 consecutive epochs

For CNN classifier, we use a one-layer CNN encoder with a linear classifier. The embedding is initialized with the 300-dimensional pretrained GloVe word embedding \cite{pennington-etal-2014-glove}. The CNN layer has 256 kernels and the size of the kernels is 3. We use max-pooling and AdamW with an initial learning rate 1e-3, batch size 32, with no warmup steps. The maximum number of epochs is 40 with early stopping after 3 consecutive non-improving epochs.

\subsection{Interpretation methods}
\label{sec:appendix_interpret}
For LIME,  Saliency Map, Integrated Gradients and DeepLift, we apply the implementation in Captum \footnote{https://github.com/pytorch/captum}. For Word Omission, we use our own implementation. 

\end{document}